\documentclass[conference]{IEEEtran}
\IEEEoverridecommandlockouts
\usepackage{cite}
\usepackage{amsmath,amssymb,amsfonts}
\usepackage{algorithmic}
\usepackage{graphicx}
\usepackage{textcomp}
\usepackage{xcolor}
\usepackage{soul}
\usepackage{float}
\usepackage{booktabs}
\usepackage{multirow}
\usepackage{multicol}
\usepackage{etoolbox}
\usepackage{hyperref}
\usepackage{placeins} 

\makeatletter
\patchcmd{\@makecaption}
  {\scshape}
  {}
  {}
  {}
\makeatletter
\patchcmd{\@makecaption}
  {\\}
  {.\ }
  {}
  {}
\makeatother
\def\tablename{Table}
\renewcommand{\figurename}{Figure}

\begin{document}

\title{Generative LLM Powered Conversational AI Application for Personalized Risk Assessment: A Case Study in COVID-19
}

 \author{
 \IEEEauthorblockN{Mohammad Amin Roshani$^{1}$, Xiangyu Zhou$^{1}$, Yao Qiang$^{1}$, \\Srinivasan Suresh$^{2}$, Steve Hicks$^{3}$, Usha Sethuraman$^{4}$, Dongxiao Zhu$^{1}$}
 \IEEEauthorblockA{1 Department of Computer Science, Wayne State University, Michigan, USA}
 \IEEEauthorblockA{2 Department of Pediatrics, UPMC Children’s Hospital of Pittsburgh, Pennsylvania, USA}
 \IEEEauthorblockA{3 Department of Pediatrics, Penn State College
 of Medicine, Pennsylvania, USA}
 \IEEEauthorblockA{4 Division of Emergency Medicine, Department of Pediatrics, Children’s
 Hospital of Michigan, Michigan, USA}

}


\maketitle

\begin{abstract}
Large language models (LLMs) have shown remarkable capabilities in various natural language tasks and are increasingly being applied in healthcare domains. This work demonstrates a new LLM-powered disease risk assessment approach via streaming human-AI conversation, eliminating the need for programming required by traditional machine learning approaches. In a COVID-19 severity risk assessment case study, we fine-tune pre-trained generative LLMs (e.g., Llama2-7b and Flan-t5-xl) using a few shots of natural language examples, comparing their performance with traditional classifiers (i.e., Logistic Regression, XGBoost, Random Forest) that are trained \textit{de novo} using tabular data across various experimental settings. We develop a mobile application that uses these fine-tuned LLMs as its generative AI (GenAI) core to facilitate real-time interaction between clinicians and patients, providing no-code risk assessment through conversational interfaces. This integration not only allows for the use of streaming Questions and Answers (QA) as inputs but also offers personalized feature importance analysis derived from the LLM's attention layers, enhancing the interpretability of risk assessments. By achieving high Area Under the Curve (AUC) scores with a limited number of fine-tuning samples, our results demonstrate the potential of generative LLMs to outperform discriminative classification methods in low-data regimes, highlighting their real-world adaptability and effectiveness. This work aims to fill the existing gap in leveraging generative LLMs for interactive no-code risk assessment and to encourage further research in this emerging field.

\end{abstract}


\section{Introduction}

\FloatBarrier

\begin{figure*}[t!]
\centering
\includegraphics[width=0.98\textwidth]{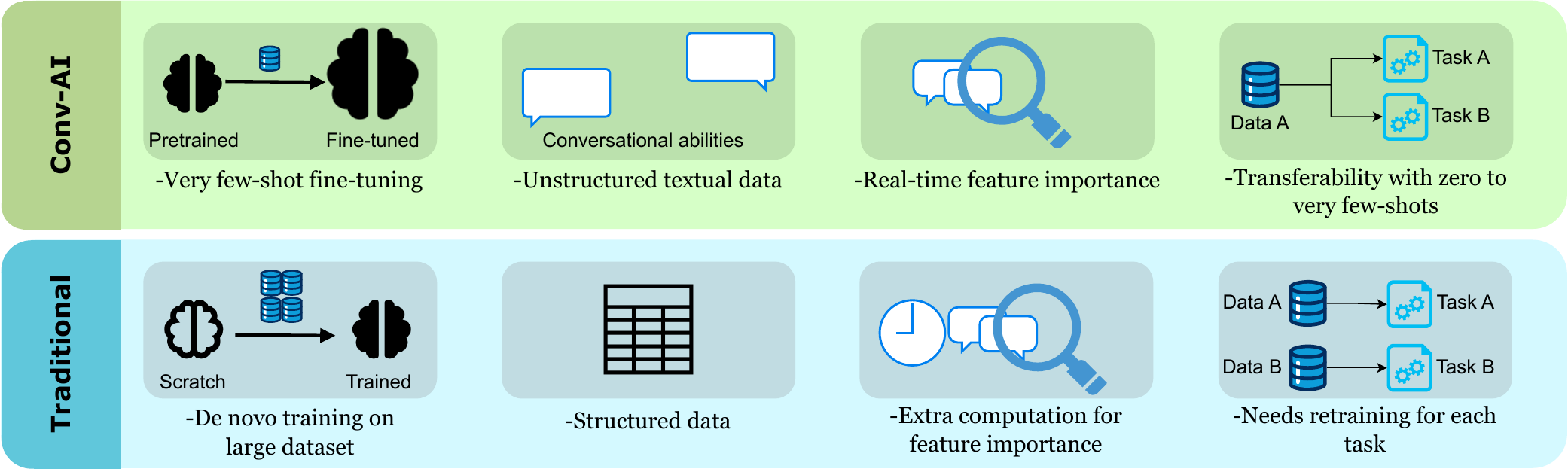}
\caption{\footnotesize A comparison between LLM-based conversational AI (Conv-AI) and traditional machine learning methods for disease risk assessment. The Conv-AI leverages pretrained models that require only very few-shot fine-tuning, can handle unstructured textual data, provide real-time feature importance for each risk assessment it provides, and offer transferability with zero to very few-shots for new risk assessment tasks. In contrast, traditional machine learning methods require large datasets for {\it de novo} training, process structured data, rely on extra computational steps for instance-specific post-hoc feature importance (e.g., SHAP), and need retraining for each new task.}

\label{fig:chat-traditional}
\end{figure*}

Disease risk assessment is a critical tool in public health surveillance, where demographic variables and social determinants are often utilized to assess a patient's susceptibility to disease, predict treatment response, and forecast severity outcomes. These predictions have been carried out using traditional classification models that are trained \textit{de novo} for each disease or condition using curated tabular data \cite{li2018leveraging,wang2019prioritization,li2020predicting}. For example, Wang et al. \cite{wang2019prioritization} developed a linear model-based multi-task learning approach to predict the risk of childhood obesity according to their geolocations. Li et al. \cite{li2020predicting} developed a mixture neural network approach to stratify patients and predict heart failure risk within each group.

The advent of transformers has marked a significant shift, allowing researchers to deploy these advanced models for various tasks, thereby improving prediction accuracy and handling complex data structures more effectively. Researchers have extensively used BERT-style models \cite{devlin2018bert} in various healthcare tasks. Notable examples include ClinicalBERT \cite{huang2019clinicalbert} and BioClinicalBERT \cite{alsentzer2019publicly}, both trained on clinical notes in the MIMIC-III database. Additionally, MedBERT \cite{rasmy2021med} was further trained on electronic health records (EHRs), resulting in high Area Under the Curve (AUC) scores for disease risk prediction. However, BERT-based models, primarily used for \textit{discriminative} tasks, are limited in their ability to process streaming question and answer (QA) pairs, such as in conversational data science tasks, due to their architecture.


\textit{Generative} LLMs, such as OpenAI's GPT-3 \cite{brown2020language}, have introduced significant advancements in Natural Language Processing (NLP) for healthcare by transcending the limitations of discriminative models like BERT. Unlike BERT-style models, which often require extensive preprocessing and are primarily tailored for specific tasks with structured inputs, generative LLMs excel at handling diverse data formats, including both structured clinical data and unstructured text such as patient narratives and medical histories. This versatility allows them to integrate and synthesize information from multiple sources, making them highly effective for complex tasks such as predicting disease severity.


With increasingly longer context windows, up to 8,192 tokens in OpenAI's GPT-4 \cite{achiam2023gpt}, generative LLMs can efficiently manage extensive patient records and interaction histories. This capability to process long, streaming, and varied inputs, coupled with their extensive pre-training on diverse datasets, allows generative LLMs to generalize effectively even with limited labeled domain-specific data. Furthermore, their ability to handle multi-hop questions and answers positions them uniquely for real-time conversational applications, facilitating no-code disease assessment via interactive patient engagements. These strengths make generative LLMs particularly suitable for tasks such as disease severity risk assessment, where leveraging pre-trained world knowledge and user-provided natural language inputs allows for accurate predictions without the need for coding.


Despite the remarkable performance of proprietary black-box LLMs, such as GPT-4 \cite{nori2023capabilities} and MedPaLM-2 \cite{singhal2023towards}, researchers are increasingly interested in deploying white-box models in healthcare and other high-stakes domains since these models can mitigate risks related to data privacy breaches and hallucination. Their transparency allows for task-specific and domain-specific fine-tuning at a reduced cost, providing researchers with complete control over the process. This shift towards encoder-decoder and decoder-only models is exemplified by PMC-LLaMA \cite{wu2023pmcllama}, a general-purpose LLM adapted from LLaMA and fine-tuned using instruction tuning on health and medical corpora, which has outperformed LLaMA-2-70B and ChatGPT-175B in several health/medical Question-and-Answer (QA) benchmarks. 

Despite these advancements, there remains a notable gap in research regarding the use of generative LLMs for disease diagnosis and risk assessment tasks. Addressing this gap is crucial for fully leveraging the potential of LLMs in healthcare applications, as they offer advanced capabilities in handling complex medical data and providing accurate predictions. One of the few studies in this area is CPLLM \cite{shoham2023cpllm}, which fine-tunes Llama2 \cite{touvron2023llama} as a general LLM and BioMedLM \cite{venigalla2022biomedlm}, trained on biological and clinical text, for different prediction tasks. Our work, however, opens a new avenue of research in conversational data science to enable no-code personalized risk assessment via a conversational interface \textit{anytime and anywhere}. We experiment with a broader range of white-box LLMs, including LLaMA2, Flan-T5, and T0 models, integrating them into a conversational agent mobile application with a natural language interface for no-code personalized risk assessment and patient-clinician communication. A comparison of our work to traditional methods is shown in Figure\ref{fig:chat-traditional}.


\begin{figure*}[t!]
\centering
\includegraphics[width=0.98\textwidth]{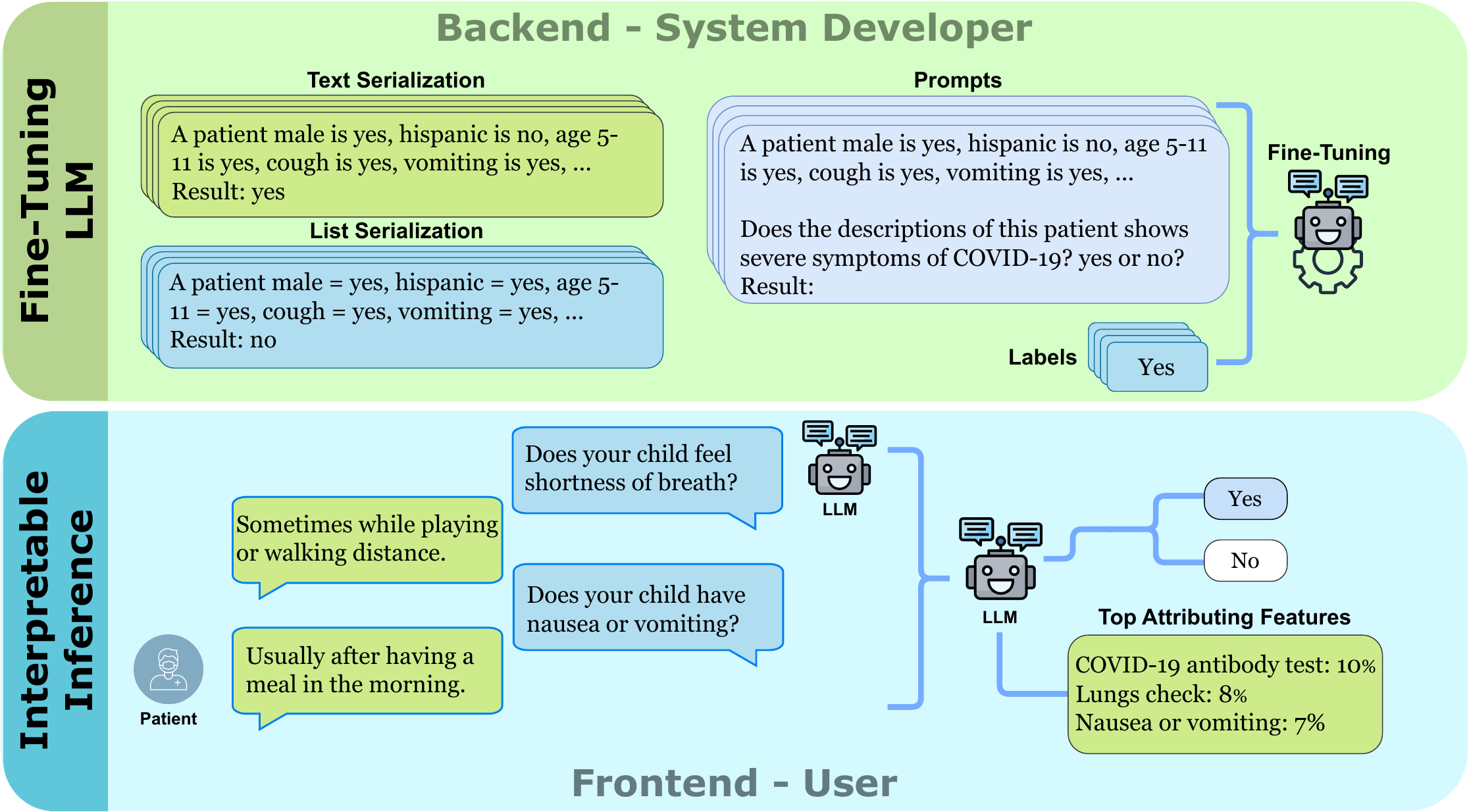}
\caption{\footnotesize Workflow for few-shot COVID-19 severity risk assessment using generative LLMs with different serialization techniques. The top section, labeled \textbf{Backend - System Developer}, shows the fine-tuning phase where a few-shot sample of patient data, serialized via List and Text Templates, is used to fine-tune the LLMs. This backend process includes the creation of prompts and corresponding labels for model fine-tuning. The bottom section, labeled \textbf{Frontend - User}, illustrates how a conversational chatbot interacts with users through our application to gather responses via streaming QAs. These responses are analyzed by the fine-tuned LLM in real-time, providing risk assessments and highlighting the top attributing features that explain the model’s risk assessment.}

\label{fig:main-method}
\end{figure*}

Our contributions to the field of LLM-based disease risk assessment are multifaceted. First and foremost, we propose a paradigm shift from traditional machine learning-based health outcome prediction, which typically relies on structured tabular data, to conversational agent-based no-code prediction using streaming QAs. This is realized through the development of a GenAI-powered mobile application that integrates fine-tuned LLMs as the core for personalized risk assessment and patient-clinician communication. The application not only assesses disease risk for patients but also provides contextual insights related to risk surveillance and mitigation through natural language conversation.

Secondly, we demonstrate that generative LLMs can outperform traditional machine learning methods (Table \ref{tab:performance}), such as Logistic Regression \cite{hosmer2013applied}, Random Forest \cite{breiman2001random}, and XGBoost \cite{chen2016xgboost}, in \textbf{low-data regimes}, which is critical for medical applications where labeled data is scarce. For instance, our results show that LLMs like the T0-3b model achieve an AUC of 0.75 in zero-shot settings, underscoring the ability of pre-trained LLMs to achieve high accuracy without task-specific training. Additionally, we provide a comprehensive comparison of both decoder-only and encoder-decoder models, fine-tuned using the widely adopted parameter-efficient LoRA (Low-Rank Adaptation) method \cite{hu2021lora}.

Thirdly, we introduce a feature importance analysis derived from the LLM's attention layers (Section~\ref{subsec:feature_importance}), providing personalized insights into the most influential factors driving the model's predictions. This enhances the interpretability and utility of the risk assessment for both patients and clinicians, offering real-time, instance-specific explanations during inference.


\section{Methods}

\subsection{Our Research Objective}
The primary objective of this research is to explore the effectiveness of pre-trained generative LLMs in no-code risk assessment of disease severity using few-shot multi-hop QAs. We aim to evaluate how these generative LLM-powered conversational agents can utilize streaming QAs to accurately classify patient outcomes as severe or non-severe, which is crucial for early risk assessment and optimizing healthcare resource allocation. Through a case study of COVID-19 severity risk assessment, we develop an application that employs open-source generative LLMs to determine the severity of COVID-19 outcomes. This involves leveraging the models' capabilities in zero-shot and few-shot settings, with a focus on the use of serialization techniques to enhance their effectiveness and generalizability. We also integrate real-time feature importance to provide interpretable risk assessments. The workflow of our approach, from fine-tuning generative LLMs using serialized QA pairs to real-time risk assessment via a conversational interface, is illustrated in Figure \ref{fig:main-method}.

\subsection{Data Collection}

A dataset was collected from the emergency departments (EDs) of Children's Hospital of Michigan and UPMC Children's Hospital of Pittsburgh between March 2021 and February 2022. The dataset includes \(n = 393\) participant records, each characterized by responses to a series of carefully designed questions. See Figure \ref{fig:application-screens} for sample QAs. The severity of outcomes was defined as the need for supplemental oxygen (\(\geq 50\%\) FiO2), non-invasive positive pressure or mechanical ventilation, extracorporeal membrane oxygenation, vasopressors or inotropes, cardiopulmonary resuscitation, or death from a related cause during hospitalization or within one month after discharge. These outcomes, categorized as severe or non-severe, were determined through chart reviews and parent surveys conducted thirty days post-discharge \cite{hicks2023saliva}.




\subsection{Tabular Data for Traditional Models}
As traditional machine learning methods require tabular data as input, we formalize the questionnaire QA pairs as $\mathcal{D} = \{(\mathbf{x}_i, y_i)\}_{i=1}^{n}$, where $n = 393$. $\mathbf{x}_i \in \{0,1\}^{d}$ represents the binary feature vector of the $i$-th instance where $d = 15$, and $y_i \in \{0,1\}$ denotes the binary class label indicating the presence or absence of severe COVID-19 symptoms determined by clinicians.

Each feature vector $\mathbf{x}_i$ consists of binary indicators representing social determinants, clinical, and demographic factors that may influence the severity of COVID-19, such as age, pre-existing conditions, vital signs, and laboratory test results. The feature names are denoted as $\mathcal{F} = \{f_1, f_2, \dots, f_{d}\}$, where each $f_j$ is a natural-language string describing the corresponding attribute.

The task is to predict the binary outcome $y_i$ based on the information provided in $\mathbf{x}_i$. This constitutes a supervised learning problem where the objective is to train a model to minimize prediction error on unseen data.

\subsection{Serialization for New Conversational AI}

At the time of data collection during 2021-2022, we did not yet have a conversational agent (chatbot) for automated data donation from users, so we used a questionnaire to collect answers from each patient based on a set of questions designed for this study. As a result, the native format of the dataset consists of QA pairs, which were subsequently serialized to fine-tune the generative LLMs for the risk assessment task. It is important to note that the fine-tuned model is capable of assessing risk using streaming QAs in real time (Figures \ref{fig:main-method} and \ref{fig:application-screens}).

To achieve serialization, the features in our dataset are denoted as $f_1, f_2, \dots, f_d$, and their associated values as $v_1, v_2, \dots, v_d$. This notation provides a structure that is transformed into natural language prompts for the LLM.

We used two main serialization methods, the \textbf{List Template} and the \textbf{Text Template}, to create natural language representations of the data. As shown in Figure \ref{fig:main-method}, the List Template links each feature with its value using an equal sign (`='), while the Text Template uses a narrative structure with the word ``is" to connect each feature with its value. These templates enable us to evaluate which serialization approach better translates the data into actionable insights by the LLM.

\FloatBarrier


\subsection{Generative LLMs}
We explore the capabilities of three white-box LLMs---LLaMA2 \cite{touvron2023llama}, T0 \cite{sanh2021multitask}, and Flan-T5 \cite{chung2024scaling}---focusing on their application in risk prediction for COVID-19 using both the native QA pairs and the formatted tabular dataset.
 To our knowledge, this is \textbf{one of the the first attempts} leveraging generative LLMs and conversational data science for disease risk assessment across various LLMs and few-shot settings. Our selection includes both decoder-only (LLaMA2) and encoder-decoder architectures (T0 and Flan-T5), allowing for a comprehensive assessment and comparison of their performance. The white-box nature of these models is particularly advantageous as it enables setup on local hosts with private datasets, ensuring precise risk assessment by allowing direct access to model weights and logits.

The input to the LLMs is a serialized string generated from the tabular data using the previously explained serialization strategies. Given a feature vector $\mathbf{x}_i = [f_1, f_2, \dots, f_d]$ and their associated values $[v_1, v_2, \dots, v_d]$, the serialized input string $S_i$ can be represented using either the List Template or Text Template serialization methods (Figure \ref{fig:main-method}).

The LLM processes the serialized input string $S_i$ and outputs logits for the next token in the sequence. We focus on the logits corresponding to the tokens `yes' and `no', which indicate severe or non-severe symptoms respectively. The probabilities for these tokens are obtained by applying the softmax function to the logits:
\[
p(\text{yes}|S_i) = \frac{e^{\text{logits}_{\text{yes}}}}{e^{\text{logits}_{\text{yes}}} + e^{\text{logits}_{\text{no}}}}
\]

The probability \( p(\text{yes}|S_i) \) indicates the likelihood of severe symptoms based on the input data \( S_i \). This probability is directly used as the severity risk score for evaluation purposes.

To determine the binary predicted label \( \hat{y}_i \) from this probability:
\[
\hat{y}_i = \begin{cases} 
1 & \text{if } p(\text{yes}|S_i) > 0.5 \\
0 & \text{otherwise}
\end{cases}
\]

The probability score \( p(\text{yes}|S_i) \), reflecting the severity risk, is used to compute the AUC for evaluation (Figure~\ref{fig:main-method}).

\subsection{Evaluation Setting}

\paragraph{Zero-Shot Setting}
In the zero-shot setting, our approach leverages the intrinsic capabilities of LLMs. These models, unlike traditional classifiers such as Logistic Regression and XGBoost, have been extensively pre-trained on diverse datasets. This extensive pre-training enables them to apply their accumulated world knowledge directly to specific classification tasks without additional training, demonstrating exceptional generalizability.

We assess the zero-shot prediction effectiveness of these LLMs by presenting them with tasks aligned with our study's objectives that they have not been specifically trained on. The models interpret and classify new, unseen data solely based on their pre-trained knowledge. This approach not only highlights the potential of LLMs in real-world applications but also evaluates their ability to generalize from their training to novel scenarios in healthcare.

This zero-shot methodology allows us to evaluate how well these LLMs can recognize and classify complex, previously unseen patterns in healthcare data, providing valuable insights into their practical applicability and limitations in clinical settings.

\paragraph{Few-Shot Fine-Tuning}
In the few-shot setting, we utilize sample sizes of 2, 4, 8, 16, and 32 to fine-tune the LLMs, aiming to examine the effect of training sample size on model performance compared to traditional classifiers. To ensure fairness and reduce bias in the fine-tuning process, we maintain a balanced ratio of positive ($y_i = 1$) and negative ($y_i = 0$) samples, with an equal number of examples from each class in each sample size.

To enhance computational efficiency in adapting the LLMs to our specific tasks, we employ a parameter-efficient fine-tuning approach using LoRA (Low-Rank Adaptation) \cite{hu2021lora}. Instead of adjusting all parameters within the model, LoRA involves training a small proportion of parameters by integrating trainable low-rank matrices into each layer of the pre-trained model. This method allows the model to quickly adapt to new tasks by optimizing only a subset of parameters, thereby preserving the general capabilities of the LLM while enhancing its performance on task-specific features.

\subsection{Feature Importance Analysis}
\label{subsec:feature_importance}

\begin{figure*}[t!]
  \centering
  \centerline{\includegraphics[width=0.8\textwidth]{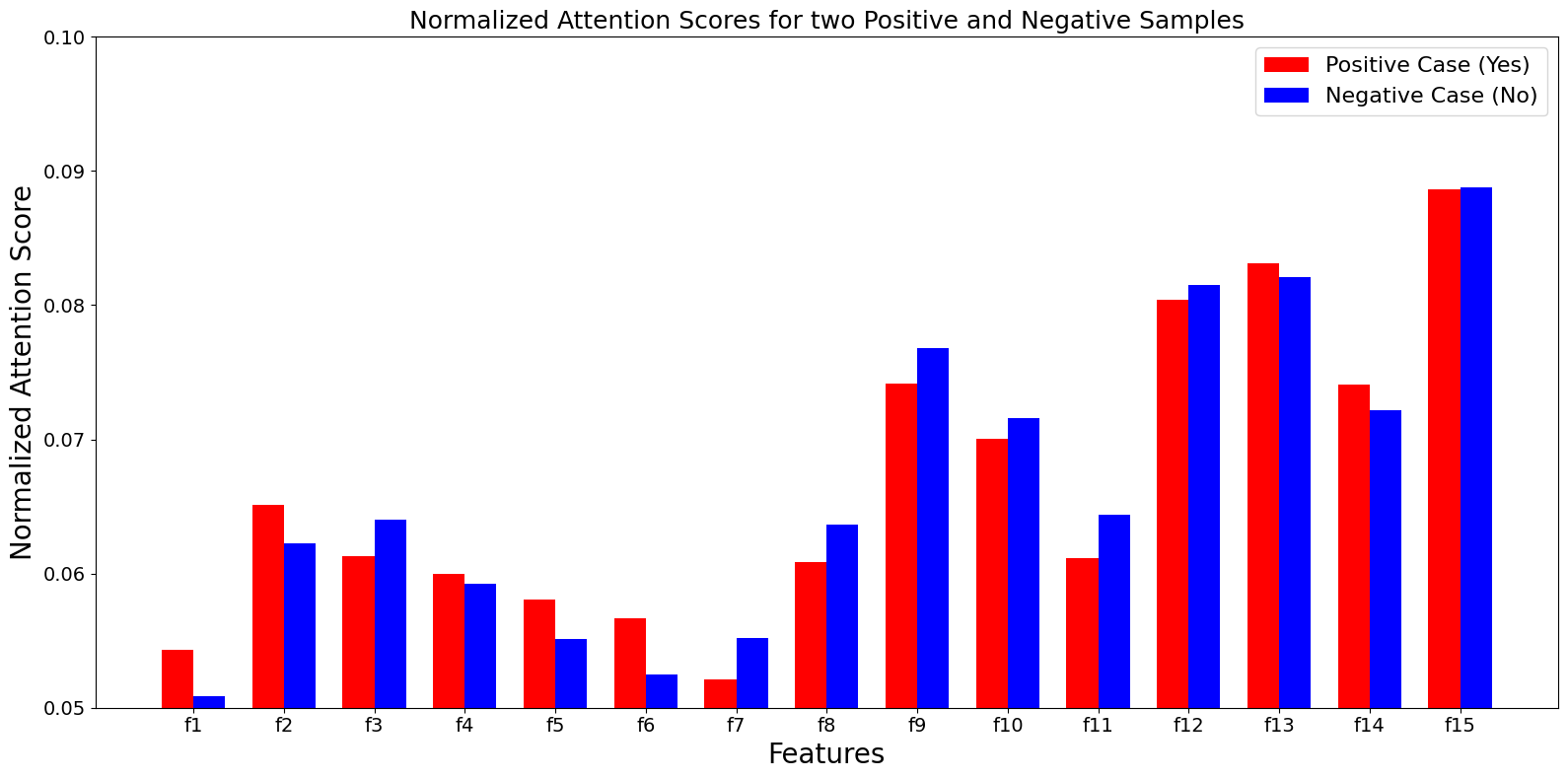}}
  \vspace{-0.2cm}
  \caption{Normalized attention scores from LLaMA2-7b in the 32-shot setting, showing feature importance for two test cases, one positive (yes) and one negative (no), {\bf simultaneously} with the risk assessment.}
  \vspace{-0.4cm}
  \label{fig:attention-scores}
\end{figure*}

In disease risk assessment, interpretability is as critical as accuracy, particularly when both are provided to the user in real-time. Here, we introduce a novel approach for analyzing feature importance by leveraging the attention mechanisms inherent in the output layers of generative LLMs. This method provides additional insights into the risk assessment process of the model, which is valuable for both clinicians and patients in understanding the factors contributing to the model's ouput.

Our approach involves extracting attention scores from the model's output layer, where the attention assigned to each input token is interpreted as an indicator of feature importance. We compute the attention for each feature-value pair and associate the average attention score with the corresponding feature. This provides a holistic view of which features, along with their associated values, influence the model’s output.

For an input sequence such as:
\begin{quote}
A patient with $f_1 = v_1$, $f_2 = v_2$, \dots, $f_{15} = v_{15}$.
Does this patient have COVID-19, yes or no?
\end{quote}

We calculate attention scores for each feature-value pair in the original sequence. The average attention score for each feature-value pair is then computed, and the score is associated with the feature itself, offering a representation of feature importance in the context of disease severity risk. 

This normalized attention score serves as a proxy for feature importance, offering clinicians and patients a clearer understanding of which features (e.g., age, pre-existing conditions, vital signs, etc.) are most influential in the model's assessment of COVID-19 severity risk. As illustrated in Figure \ref{fig:attention-scores}, the plot shows the normalized attention scores from the LLaMA2-7b model in the 32-shot setting for two test cases: one positive (yes) and one negative (no). 

For the positive case, the top five features with the highest attention scores, as shown in this figure, are:

\begin{enumerate}
    \item \textbf{f15:} COVID-19 antibody test
    \item \textbf{f13:} Lungs check
    \item \textbf{f12:} Nausea or vomiting
    \item \textbf{f9:} Cough
    \item \textbf{f14:} Eye redness
\end{enumerate}

By integrating this analysis into our mobile application, we enhance the interpretability of LLM-based risk assessments, empowering users with deeper insights into the model's reasoning process.

\begin{figure*}[t!]
\centering
\centerline{\includegraphics[width=0.85\textwidth]{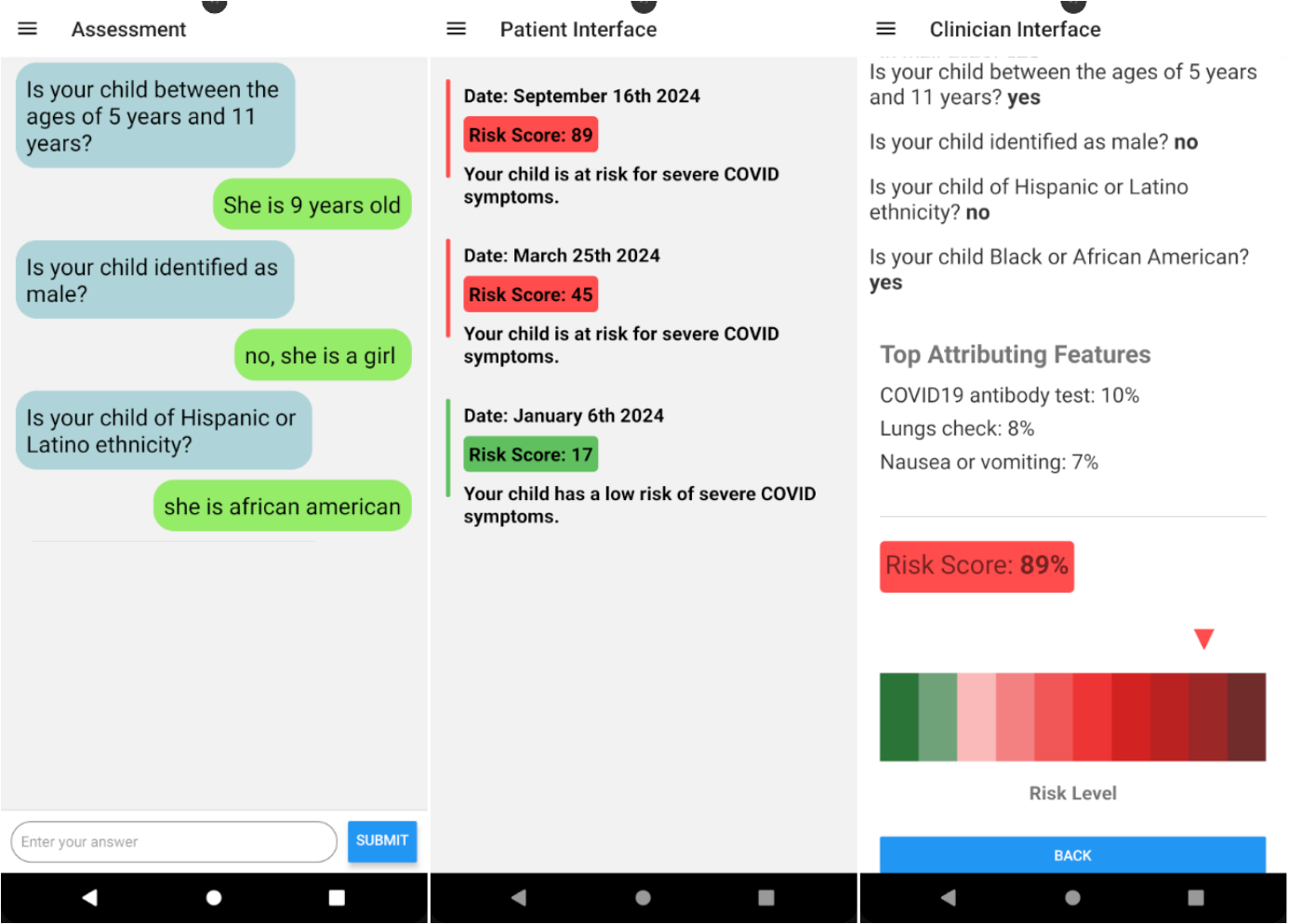}}
\caption{\footnotesize Overview of our mobile application design, showcasing patient data collection, real-time risk assessment using LLMs, and clinician review interface.}
\label{fig:application-screens}
\end{figure*}

\section{Mobile Application}

To provide users with code-free disease severity risk assessment and enhance user experience, we developed a mobile conversational agent powered by the aforementioned generative LLMs. This application is designed to facilitate the assessment and management of COVID-19 in children, with potential applicability to other diseases and conditions. It offers two versions: one for patients to donate their health information via answering the questions and receive real-time severity risk assessments, and another for clinicians to manage, review, and interpret the sessions donated by patients. The primary goals are to enhance early detection of severe outcomes, improve patient-clinician communication, and streamline the overall risk assessment process.

The application targets patients, clinicians, and other healthcare providers involved in managing pre-clinical cases. It leverages the capabilities of generative LLMs to analyze patient responses and provide immediate feedback on the risk of severe symptoms. Developed using React Native and JavaScript for the front end, Firebase for database management, and various frontend technologies, the application provides a user-friendly, efficient, and effective solution for managing disease risks. It aims to improve patient outcomes by facilitating timely and informed decision-making.

\subsection{Database Structure}
Our mobile application utilizes Firebase for database management, structured into three primary collections: Users, Questions, and Answers.

\begin{itemize}
    \item \textbf{Users}: This collection includes essential user information such as \texttt{ID}, \texttt{Email}, and \texttt{isAdmin}. The \texttt{ID} uniquely identifies each user, the \texttt{Email} serves as contact information, and the \texttt{isAdmin} field (a boolean) indicates whether the user has administrative privileges (clinicians) or not (patients).

    \item \textbf{Questions}: Each document in this collection has a unique \texttt{ID} and a \texttt{Description} field. The \texttt{ID} is used to reference questions in the \texttt{Answers} collection, and the \texttt{Description} contains the text of the question posed to the user, ensuring clarity and specificity in data mapping.

    \item \textbf{Answers}: This collection records user responses during their sessions. Each document includes a session \texttt{ID}, an array of \texttt{Answers} where each entry links to the relevant \texttt{QuestionID} from the \texttt{Questions} collection. Additionally, it contains a \texttt{Text} field for the user's detailed response, an \texttt{Answer} field for the LLM-generated response (e.g., \texttt{Yes} or \texttt{No}), a \texttt{Date} field marking the session's completion time, a \texttt{Risk Score} field, which is derived from the user’s responses and utilized for subsequent risk prediction by the LLM, and an \texttt{Important Features} field, which stores the key features identified by the LLM’s attention scores that contributed to the risk assessment.
\end{itemize}

\begin{table*}[ht!]
\scriptsize
\centering
\caption{Performance of models across different shot settings. All values represent the AUC rounded to two decimal places. Standard deviations given across five random seeds are shown as subscripts. The suffixes -L and -T represent List Serialization and Text Serialization, respectively.}
\label{tab:performance}
\resizebox{0.75\textwidth}{!}{%
\begin{tabular}{@{}lcccccc@{}}
\toprule
\textbf{Model} & \multicolumn{6}{c}{\textbf{Number of Shots}} \\ 
\cmidrule(l){2-7} 
& \textbf{0} & \textbf{2} & \textbf{4} & \textbf{8} & \textbf{16} & \textbf{32} \\ \midrule
Llama2-7b-L & $0.54_{.05}$ & $0.69_{.07}$ & $0.69_{.06}$ & $0.68_{.04}$ & $0.63_{.04}$ & $0.66_{.07}$ \\
Flan-t5-xl-L & $0.62_{.03}$ & $0.64_{.04}$ & $0.63_{.02}$ & $0.68_{.06}$ & $0.66_{.05}$ & $0.69_{.06}$ \\
Flan-t5-xxl-L & $0.60_{.03}$ & $0.61_{.03}$ & $0.61_{.05}$ & $0.62_{.06}$ & $0.59_{.10}$ & $0.65_{.11}$ \\
T0pp(8bit)-L & $0.69_{.04}$ & $\mathbf{0.70_{.07}}$ & $\mathbf{0.70_{.05}}$ & $0.70_{.05}$ & $0.68_{.06}$ & $\mathbf{0.70_{.10}}$ \\
T0-3b-L & $0.68_{.04}$ & $0.67_{.04}$ & $0.68_{.05}$ & $0.70_{.04}$ & $0.67_{.04}$ & $0.67_{.07}$ \\ 
\midrule
Llama2-7b-T & $0.59_{.05}$ & $0.69_{.03}$ & $0.69_{.01}$ & $0.64_{.07}$ & $0.63_{.05}$& $0.67_{.06}$ \\
Flan-t5-xl-T & $0.69_{.03}$ & $0.69_{.02}$ & $0.69_{.03}$ & $\mathbf{0.71_{.05}}$& $\mathbf{0.69_{.04}}$& $\mathbf{0.70_{.05}}$ \\
Flan-t5-xxl-T & $0.61_{.04}$ & $0.58_{.03}$& $0.63_{.08}$& $0.59_{.10}$& $0.62_{.09}$ & $0.63_{.10}$ \\
T0pp(8bit)-T & $0.67_{.02}$ & $0.65_{.05}$ & $0.66_{.05}$ & $0.68_{.04}$ & $0.65_{.08}$& $0.67_{.08}$ \\
T0-3b-T & $\mathbf{0.75_{.04}}$ & $0.65_{.06}$ & $0.65_{.05}$ & $0.68_{.03}$ & $0.67_{.04}$ & $0.65_{.08}$ \\ 
\midrule
Logistic Regression & $-$ & $0.57_{.07}$ & $0.55_{.10}$ & $0.64_{.06}$ & $0.61_{.11}$ & $0.69_{.08}$ \\
Random Forest & $-$ & $0.57_{.07}$ & $0.57_{.06}$ & $0.62_{.08}$ & $0.66_{.07}$ & $0.68_{.07}$ \\
XGBoost & $-$ & $0.50_{.00}$ & $0.50_{.00}$ & $0.50_{.00}$ & $0.54_{.06}$ & $0.65_{.03}$ \\
\bottomrule
\end{tabular}%
}
\end{table*}

\subsection{User Interface - Assessment}
As shown in Figure \ref{fig:application-screens}, on the \textbf{Assessment} page, we leverage the power of LLMs to engage in a conversation with the patient. This interaction allows us to ask questions and gather contextual information for each response. By doing so, we retrieve a binary answer (Yes/No) using the LLM, which is then provided to the primary care physician along with the patient's context to aid in decision-making.


After the user responds to each question, we use our LLM to generate a binary answer. This involves providing the LLM with an instruction that includes the question and the user's response, asking the LLM to interpret the response into a binary answer (Yes or No). This sequential process is performed for all questions. Currently, the input for the final LLM-based risk assessment, which predicts the COVID-19 severity risk, is based solely on the set of binary answers generated by the LLM. Future enhancements could incorporate the original user responses to improve context understanding.

We currently utilize the Llama2-7b API for answer retrieval. Our long-term goal is to integrate a fine-tuned LLM hosted on our servers to ensure better optimization and accuracy specific to our dataset, as evidenced by the improved performance results discussed in this paper.

\subsection{User Interface - Patient and Clinician Results}
Patients can submit a session at any time, receiving an immediate risk assessment in the \textbf{Patient Results} section (see Figure \ref{fig:application-screens}). This section displays all sessions submitted by the current user, along with their respective risk assessments.

In the \textbf{Clinician Results} section, clinicians can access all sessions from their patients, organized by patient ID for efficient review. Each session includes a comprehensive report featuring the predicted risk score, ensuring transparency and aiding in clinical decision-making.

Upon submission, a patient's session is instantly available in both the patient’s and clinician’s panels. While patients can only view their own sessions, clinicians can review all sessions from their assigned patients. This setup supports real-time updates through Firebase, facilitating seamless communication and follow-up between patients and their healthcare providers. Moreover, the application provides personalized feature importance analysis based on the LLM's attention layers, giving both patients and clinicians additional insights into the most critical factors influencing the risk assessment.

\section{Experimental Results}

\subsection{Training and Fine-Tuning Settings}
In our experiments, we employed a rigorous setup using five specific random seeds—0, 1, 32, 42, and 1024—to ensure diverse dataset initialization and mitigate potential biases in data allocation.

For traditional machine learning methods, the dataset of 393 samples was divided into 65\% training, 15\% validation, and 20\% testing segments. Although the full training set is available, we focus specifically on training the models with up to 32 shots to examine performance in the few-shot regime. For LLMs, we similarly fine-tune the models using up to 32 shots, highlighting their capability to generalize in low-data settings with minimal task-specific examples.

When fine-tuning LLMs using LoRA, we monitored the validation loss to select the best model checkpoint, aiming to minimize overfitting and enhance generalization to the test set.


\begin{figure*}[t!]
  \centering
  \centerline{\includegraphics[width=0.98\textwidth]{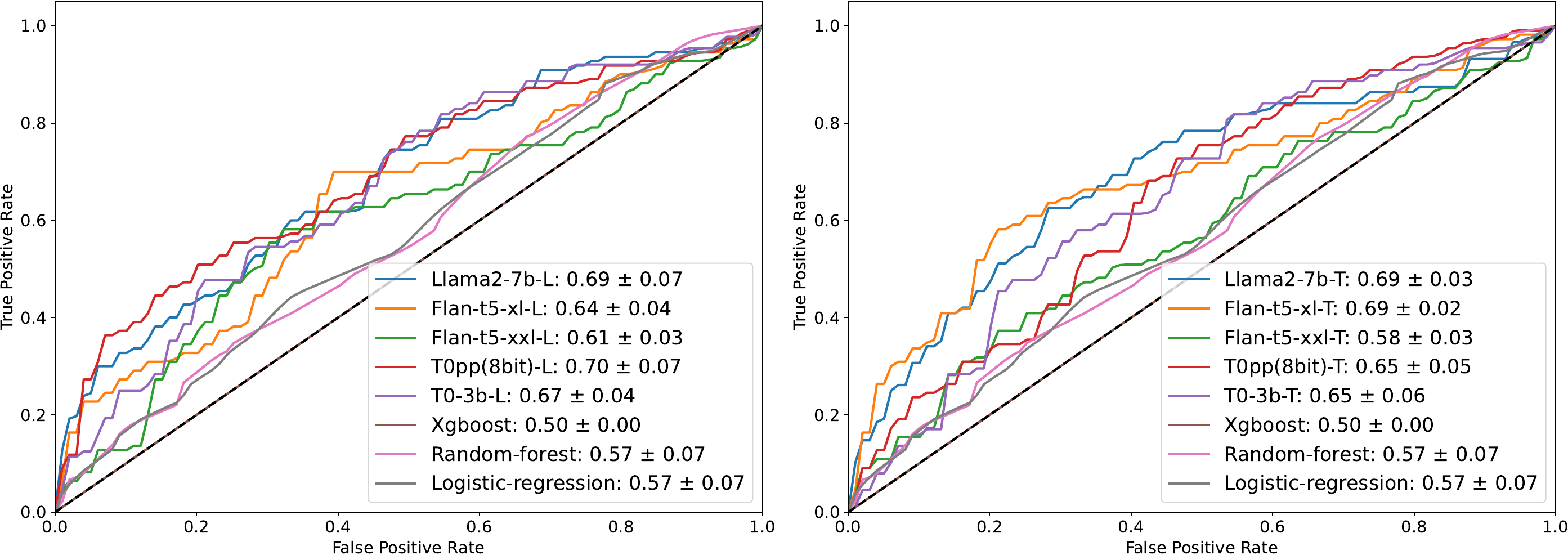}}
  \caption{Average AUC in \textbf{2-shot} setting over five different seeds. The left panel shows results using the List Serialization (-L) approach, while the right panel shows results using the Text Serialization (-T) approach.
}
  \label{fig:roc-32}
\end{figure*}

\subsection{Effects of Serialization}
Table \ref{tab:performance} shows the performance of different serialization methods for the LLMs across various few-shot settings. We evaluated two primary serialization methods: List Template and Text Template, across models tested with 0, 2, 4, 8, 16 and 32 training shots to observe performance variations with the number of training examples.

The List Template often exhibited better performance at lower shot counts, while the Text Template typically outperformed the List Template as the number of training examples increased. The following summarizes the performance trends for each model:

\begin{itemize}
    \item \textbf{Llama2-7b}: 
    In the zero-shot setting, the Text Template achieved an AUC of 0.59 compared to 0.54 for the List Template. At 2 training shots, both templates achieved an AUC of 0.69, but the Text Template began to outperform, reaching an AUC of 0.67 at 32 training shots compared to 0.66 for the List Template.

    \item \textbf{Flan-t5-xl}: 
    The Text Template consistently outperformed the List Template across most shot settings. At 2 training shots, the Text Template achieved an AUC of 0.69 compared to 0.64 for the List Template, and this lead continued up to 32 shots, where the Text Template achieved an AUC of 0.70 compared to 0.69 for the List Template.

    \item \textbf{Flan-t5-xxl}: 
    Both templates showed similar performance in the early few-shot settings. At 2 training shots, the List Template achieved an AUC of 0.61, slightly outperforming the Text Template, which achieved an AUC of 0.58. By 32 training shots, the List Template achieved an AUC of 0.65, slightly outperforming the Text Template, which achieved an AUC of 0.63.

    \item \textbf{T0pp (8bit)}: 
    In the zero-shot setting, the List Template led with an AUC of 0.69 compared to 0.67 for the Text Template. This lead was maintained through most shot settings, with both templates achieving around 0.70 AUC by 32 shots.

    \item \textbf{T0-3b}: 
    The Text Template outperformed the List Template in the zero-shot setting, achieving an AUC of 0.75 compared to 0.68 for the List Template. In the 2-shot setting, the List Template performed slightly better with an AUC of 0.67 compared to 0.65 for the Text Template. At 32 shots, the Text Template closed the gap with an AUC of 0.65 compared to 0.67 for the List Template.
\end{itemize}

Overall, while the List Template often provides an initial advantage in early few-shot settings, the Text Template shows competitive performance as the number of training examples increases. This suggests that serialization choice can be important in low-data regimes. The Text Template's strong performance in the zero-shot setting, particularly for the T0-3b model, highlights its potential when no training data is available.

\subsection{LLMs vs Traditional Machine Learning Methods}

Our study highlights the versatility of LLMs for various healthcare applications, particularly in scenarios with limited data. To benchmark their performance against traditional machine learning methods, we compared LLMs with Logistic Regression, Random Forest, and XGBoost.

LLMs benefit from extensive pre-training, allowing them to generalize well to ``unseen" data, unlike traditional methods that require substantial amounts of training data. As shown in Table \ref{tab:performance}, LLMs like T0-3b-T achieved an AUC of 0.75 in the zero-shot setting, outperforming traditional methods even without task-specific fine-tuning. This demonstrates the effectiveness of LLM-powered risk assessment without the need for additional labeled data.

In the 2-shot setting, LLMs continue to show strong performance relative to traditional methods. For instance, Figure \ref{fig:roc-32} compares the average AUC across five different seeds in this scenario. The left panel shows results using the List Serialization (-L) approach, while the right panel shows results using the Text Serialization (-T) approach. In this 2-shot scenario, LLMs such as T0pp(8bit)-L and Flan-t5-xl-T achieve AUCs of 0.70 and 0.69, respectively, clearly outperforming traditional methods, including Logistic Regression, Random Forest, and XGBoost, which achieved AUCs of 0.57, 0.57, and 0.50, respectively.

LLMs’ ability to perform well with minimal data highlights their advantage in low-data regimes. This makes them particularly suitable for real-time, no-code healthcare applications where rapid decision-making is required, even in scenarios where labeled data is scarce. 

Furthermore, LLMs' capacity to handle streaming data formats, such as multi-hop question-answering (QA), enhances their integration into conversational interfaces, supporting real-time patient-clinician interactions. This flexibility offers significant utility in clinical settings where personalized and immediate risk assessments are needed (Figure \ref{fig:chat-traditional}).

Overall, while traditional methods may improve with larger datasets, LLMs demonstrate a clear advantage in dynamic, low-data healthcare environments. Their ability to handle incomplete data and streaming input formats makes them robust for real-world applications requiring adaptability and speed.

\section{Discussion}

Our research demonstrates that generative LLMs provide a robust and no-code approach for predicting COVID-19 severity, particularly effective in low-data regimes. These models excel in zero-shot and few-shot settings, showcasing their ability to perform well without extensive domain-specific training. This is crucial for real-time applications requiring immediate and reliable predictions, highlighting their exceptional generalizability compared to traditional classifiers like Logistic Regression, Random Forest, and XGBoost, which typically require more labeled data to achieve comparable performance.

Generative LLMs effectively handle diverse input formats, integrating both structured clinical data and unstructured natural language inputs from patient interactions. This flexibility enables them to synthesize information from various sources, such as patient medical histories and symptom descriptions, enhancing their utility in dynamic healthcare settings. In our study, we incorporated these models into a conversational interface, which facilitates real-time patient-clinician interactions and immediate risk assessments. This setup supports continuous data collection and leverages the conversational capabilities of LLMs to optimize clinical decision-making and resource allocation.

Future work should focus on integrating continuous clinician-patient conversational data for fine-tuning or in-context learning (ICL), extending the application of LLMs beyond static disease prediction models. Techniques like Chain of Thought (CoT) and Chain of Interaction (CoI), which align with the interactive nature of medical consultations, show promise for enhancing model performance in interpreting and responding to patient data in real-time settings \cite{han2024chain, gramopadhye2024few}.

While our study utilized models like T0pp with parameter-efficient fine-tuning using LoRA, future research could explore newer and more advanced small language models such as LLaMA3-8b and Mistral-7b-Instruct, which have demonstrated exceptional performance in low-data regimes. These models could offer greater efficiency and accuracy as computational resources and methodologies advance, supporting more sophisticated and scalable applications in healthcare.

However, as these models continue to evolve, addressing their vulnerabilities remains critical. Studies have demonstrated that adversarial attacks can hijack LLMs during in-context learning, undermining their performance in sensitive tasks such as disease risk assessment \cite{qiang2023hijacking}. In adversarial in-context learning (ICL) scenarios, an attacker can manipulate inputs, influencing the model to produce inaccurate or harmful predictions. This poses significant risks in high-stakes settings like healthcare, where incorrect assessments could lead to adverse patient outcomes. As LLMs gain wider adoption in healthcare, enhancing their resilience against such adversarial techniques is essential to ensure safe and reliable patient outcomes.

In conclusion, generative LLMs offer a valuable tool for no-code risk assessment in low-data regimes. Their ability to perform zero-shot or few-shot transferability to new diseases or conditions and handle complex, varied inputs positions them as key assets for enhancing healthcare interventions and resource management. Furthermore, the incorporation of feature importance analysis derived from the LLM's attention layers provides an additional layer of interpretability, offering personalized insights into the decision-making process for both patients and clinicians.


\bibliographystyle{IEEEtran}
\bibliography{ref}

\end{document}